\documentclass[journal]{IEEEtran}
\usepackage{blindtext,graphicx,booktabs,amsmath,amssymb,tabularx,tabu,mathrsfs}
\usepackage[ruled]{algorithm2e}
\usepackage{stfloats}
\usepackage{color}
\usepackage{xcolor}

\makeatletter
\def\endthebibliography{%
  \def\@noitemerr{\@latex@warning{Empty `thebibliography' environment}}%
  \endlist
}
\makeatother

\newcommand{\frenet}{Frenét }

\begin{document}
\title{Identify, Estimate and Bound the Uncertainty of Reinforcement Learning for Autonomous Driving}

\author{Weitao Zhou, Zhong Cao, Nanshan Deng, Kun Jiang, Diange Yang
\thanks{All authors are with the School of Vehicle and Mobility, Tsinghua University, Beijing, China 100084.
 {\tt\small \{zwt19,dns18\}@mails.tsinghua.edu.cn;}  {\tt\small\{caozhong,jiangkun,ydg\}@tsinghua.edu.cn;}
 
Z. Cao and D. Yang are the corresponding authors.

}
}

\markboth{IEEE TRANSACTIONS ON INTELLIGENT TRANSPORTATION SYSTEMS}%
{Shell \MakeLowercase{\textit{et al.}}: Bare Demo of IEEEtran.cls for Journals}

\maketitle

\begin{abstract}
Deep reinforcement learning (DRL) has emerged as a promising approach for developing more intelligent autonomous vehicles (AVs). A typical DRL application on AVs is to train a neural network-based driving policy. However, the black-box nature of neural networks can result in unpredictable decision failures, making such AVs unreliable. 
To this end, this work proposes a method to identify and protect unreliable decisions of a DRL driving policy. The basic idea is to estimate and constrain the policy’s performance uncertainty, which quantifies potential performance drop due to insufficient training data or network fitting errors. By constraining the uncertainty, the DRL model's performance is always greater than that of a baseline policy. 
The uncertainty caused by insufficient data is estimated by the bootstrapped method. Then, the uncertainty caused by the network fitting error is estimated using an ensemble network. Finally, a baseline policy is added as the performance lower bound to avoid potential decision failures. The overall framework is called uncertainty-bound reinforcement learning (UBRL). The proposed UBRL is evaluated on DRL policies with different amounts of training data, taking an unprotected left-turn driving case as an example. The result shows that the UBRL method can identify potentially unreliable decisions of DRL policy. The UBRL guarantees to outperform baseline policy even when the DRL policy is not well-trained and has high uncertainty. Meanwhile, the performance of UBRL improves with more training data. Such a method is valuable for the DRL application on real-road driving and provides a metric to evaluate a DRL policy.

\end{abstract}


\IEEEpeerreviewmaketitle

\section{Introduction}

Autonomous driving technology is believed to bring safer and more efficient transportation in the future. While rule-based autonomous vehicle (AV) systems have demonstrated remarkable performance in specific driving tasks, they often struggle with stochastic and interactive real-world driving scenarios \cite{jain2021autonomy} \cite{wei2021autonomous}. Recently, deep reinforcement learning (DRL) methods provide a promising  route for designing smarter AV systems capable of automatically improving performance from data. These methods have achieved impressive performance in various challenging driving scenarios, e.g., highway exiting \cite{cao2020highway}, unprotected left-turn \cite{zhou2020integrating}, roundabout driving \cite{cao2021confidence} \textit{et. al.}

However, the black-box nature of DRL methods raises significant safety concerns, as the system may fail unexpectedly if the model is not properly trained. This issue is commonly referred to as the model uncertainty problem \cite{hullermeier2021aleatoric}. Notably, the model uncertainty should be distinguished from the uncertainty from the driving environment, e.g., the stochastic of agents' intention \cite{dabney2018implicit} \cite{mihatsch2002risk}, or the noise and error from the sensors \cite{rasouli2019autonomous}. 
The model uncertainty arises from the training process \cite{hullermeier2021aleatoric}. 
The training data is limited; while the reality has infinite cases, thus there always exist corner cases for a DRL model. Meanwhile, the black-box model may not learn all the knowledge from data due to a flaw in its structure. Therefore, the model uncertainty problem of DRL should be carefully considered before the trained model is applied to safe-critical AVs.

Related works to tackle DRL's model uncertainty can be divided into 1) Dangerous actions correction and 2) Training uncertainty constraints.

The dangerous actions correction method tackles the model uncertainty problem by monitoring the outputs of the DRL model during driving. Typically, a safe action set is designed to identify hazardous model outputs, e.g., using a rule-based safe-guard \cite{chen2017obstacle} or variance of policy outputs estimated by ensemble networks \cite{hoel2020reinforcement} \cite{hoel2020tactical}. In addition, the responsibility-sensitive-safety (RSS) method employs a formal model to check if the output leads to responsible collisions or accidents \cite{shalev2017formal}.  Once risks are detected,  the AV can switch to a backup policy, e.g., using decision tree methods \cite{li2017explicit}. 
These studies provide a straightforward way to protect the black-box models externally without considering the source of uncertainty. However, the pre-designed sets/thresholds may be limited to specific scenarios and could still fail unexpectedly. Simultaneously, the DRL policy might be hindered from achieving smarter performance, particularly in cases where optimal actions fall outside the safe action set.
 
The training uncertainty constraint methods  aim to limit the use of the DRL model in potentially high-uncertainty cases. 
The importance sampling method can estimate the uncertainty arising from the off-policy setting, i.e., the learned policy differs from the policy that collected training data \cite{thomas2015high}. 
{Some previous works estimate the uncertainty due to insufficient training data using Lindbergh-Levy's law \cite{cao2021confidence} \cite{cao2022trustworthy}.
Two representative deep learning techniques to estimate the uncertainty of neural networks are Bayesian networks and ensemble methods.
The Bayesian network method modifies network weights as random variables to build a distribution for network output \cite{gal2016dropout}. 
The ensemble methods build multiple parallel models to capture the uncertainty \cite{ganaie2021ensemble} \cite{zhou2022long} \cite{zhou2022dynamically}. 
The estimated uncertainty can be utilized to build the safe policy improvement process, i.e., the policy is guaranteed to be improved with more training data \cite{simao2019safe, ozisik2020security}.
Another common approach is generating a conservative DRL policy that avoids high-uncertainty states by reducing the values of such states\cite{kumar2020conservative, yu2021combo}  or encouraging the agent to act within data-covered states \cite{wu2019behavior}.
These methods improve safety during the usage of the DRL model, as the agent will be less likely to reach the high-uncertainty states that may lead to risky decisions.  However, this solution may not be practical for on-road AVs, which cannot control the encountered driving cases and thus is still risky in high-uncertain cases. }

To this end, this paper proposes a method to identify and constrain DRL uncertainty, which quantifies the potential decline in performance attributable to inadequate training data or network fitting error. The constraint is achieved using a baseline policy as the performance lower bound, i.e., the DRL model's performance should always be greater than that of the baseline policy. 
To achieve that, the DRL uncertainty is estimated by tracking the policy generation process. The DRL will only be activated when it has low uncertainty and can outperform the baseline policy. The overall framework is called uncertainty-bound reinforcement learning (UBRL). The key contributions of this work are:

1) Estimate the uncertainty due to insufficient training data: The uncertainty is estimated based on the statistics bootstrap  method. The built estimator re-samples several sub-datasets from the originally collected dataset to infer a distribution of possible policy performance, which quantifies uncertainty due to insufficient data. (See Section III.B)

2) Estimate the uncertainty caused by neural network fitting error: The uncertainty is estimated using an ensemble network that includes multiple parallel value networks with identical structures. The weights of these value networks are initialized randomly and trained separately. The outputs of multiple networks indicate uncertainty caused by neural network fitting. (See Section III.C)

3) Constraint the uncertainty: A performance lower bound is established using a baseline policy. The estimated uncertainty identifies the potential risky/low-performance decisions of DRL. Then the DRL will only be activated when it has low uncertainty and can outperform the baseline policy. This way, the DRL will be protected in risky cases without restricting its potential for high performance. (See Section III.D, E)

The remainder of this paper is organized as follows: Section II introduces the preliminaries and formally defines the problem. The uncertainty-bound reinforcement learning (UBRL) method is introduced in Section III. Section IV shows the case study and simulation results. Finally, Section V concludes this work.

\section{Preliminaries and Problem Definitions}
\subsection{Preliminaries}

Following \cite{kochenderfer2015decision}, the planning problem of AVs can be formulated as a Markov decision process (MDP) or a partially observable Markov decision process (POMDP).
The MDP assumption makes the problem satisfy the Markov property: the conditional probability distribution of future states of the process depends only on the present state. 
The agent (AV) should optimize long-term rewards under a sequential decision-making setting. 

In detail, an MDP can be defined by the tuple $(\mathcal{S}, \mathcal{A}, \mathcal{R}, \mathcal{P})$ consisting of:

\begin{itemize}
\item a state space $\mathcal{S}$
\item an action space $\mathcal{A}$
\item a reward function $\mathcal{R}$
\item a transition operator (probability distribution) $\mathcal{P}$: $\mathcal{S}\times\mathcal{A}\times \mathcal{S}$
\item a discount factor $\gamma\in(0,1]$ is set as a fixed value to favour immediate rewards over rewards in the future.
\item the planning horizon is defined as $H\in\mathbb{N}$. 
\end{itemize}

A general policy $\pi\in\Pi$ maps each state to a distribution over actions.  Here, $\Pi$ denotes the set containing all candidate policies $\pi$. 
Namely, $\pi(a|s)$ denotes the probability of taking action $a$ in state $s$ using the policy $\pi$. A policy $\pi$ has associated value $V_{\pi}$ and action-value functions $Q_{\pi}$. 
For a given reward function $\mathcal{R}$, the value function denotes expected reward for state $V(s)$ and state-action pair $Q(s, a)$ can be defined as:

\begin{equation}
      V_{\pi}(s)=\mathbb{E}_{\pi}[\sum_{t=0}^{\infty} \gamma^{t} r_t|s_0=s] \\
\label{equ:Vdefine}
\end{equation}
\begin{equation}
      Q_{\pi}(s,a)=\mathbb{E}_{\pi}[\sum_{t=0}^{\infty}  \gamma^{t}r_t|s_0=s,a_0=a]
\label{equ:Qdefine}
\end{equation}
where $a_t\thicksim \pi(a_t|s_t),s_{t+1} \thicksim P(s_{t+1}|s_t,a_t), r_t = \mathcal{R}(s_t,a_t)$. $\mathbb{E}[\cdot]$ denotes the expectation with respect to this distribution. 

Reinforcement learning (RL) planners have the potential to solve MDP problems by interacting with the environment. In each planning step, the RL agent makes decisions and outputs  action  $a$ with the state $s$ as input; then, it receives the next state $s'$ and reward $r$ from the environment. That is, an RL agent collects  data in each MDP step $\{s,a,r,s’\}$. The data will then be stored in the dataset $D$. To find the optimal  policy in the environment,  the RL agent should maximize the expected reward of the policy:

\begin{equation}
\pi^{*}(s)=\underset{\pi}{\operatorname{argmax}} \mathbb{E}_{\pi}[\sum_{t=0}^{\infty} \gamma^{t} r_t|s_0=s]
\end{equation}
To scale to large problems, deep reinforcement learning (DRL) planners often learn a parameterized estimation of the optimal Q-value function $Q_\theta\left(s, a\right) $ using a neural  network. A dataset $D$ is defined to include several trajectories collected for policy training. A trajectory started from state $s$ and collected using policy $\pi$ for $H$ future time steps can be represented as $\omega_{\pi}(s)$:

\begin{equation}
\omega_{\pi}(s):=\left\{s_0, a_0, r_0, s_1, a_1, r_1\ldots, s_{H}, a_{H}, r_{H}\right\}
\label{trajectory}
\end{equation}The parameter $\theta$ of the optimal  Q-value function $Q_\theta\left(s, a\right) $ is learned from dataset $D$ with learning function $f$. 
Then the policy $\pi_\text{rl}(s)$ will try to maximize the expected reward using the trained optimal value network:

\begin{equation}
Q_\theta\left(s, a\right) \leftarrow f(D)
\label{drllearning}
\end{equation}
\begin{equation}
\pi_\text{rl}(s)=\operatorname{argmax}_{a} Q_\theta\left(s, a\right) 
\label{maxq}
\end{equation}
In this article, we use the Deep Q-learning (DQN) \cite{mnih2015human} method to learn the DRL policy.

\begin{figure}[ht]
	\centering
	\includegraphics[width=\linewidth]{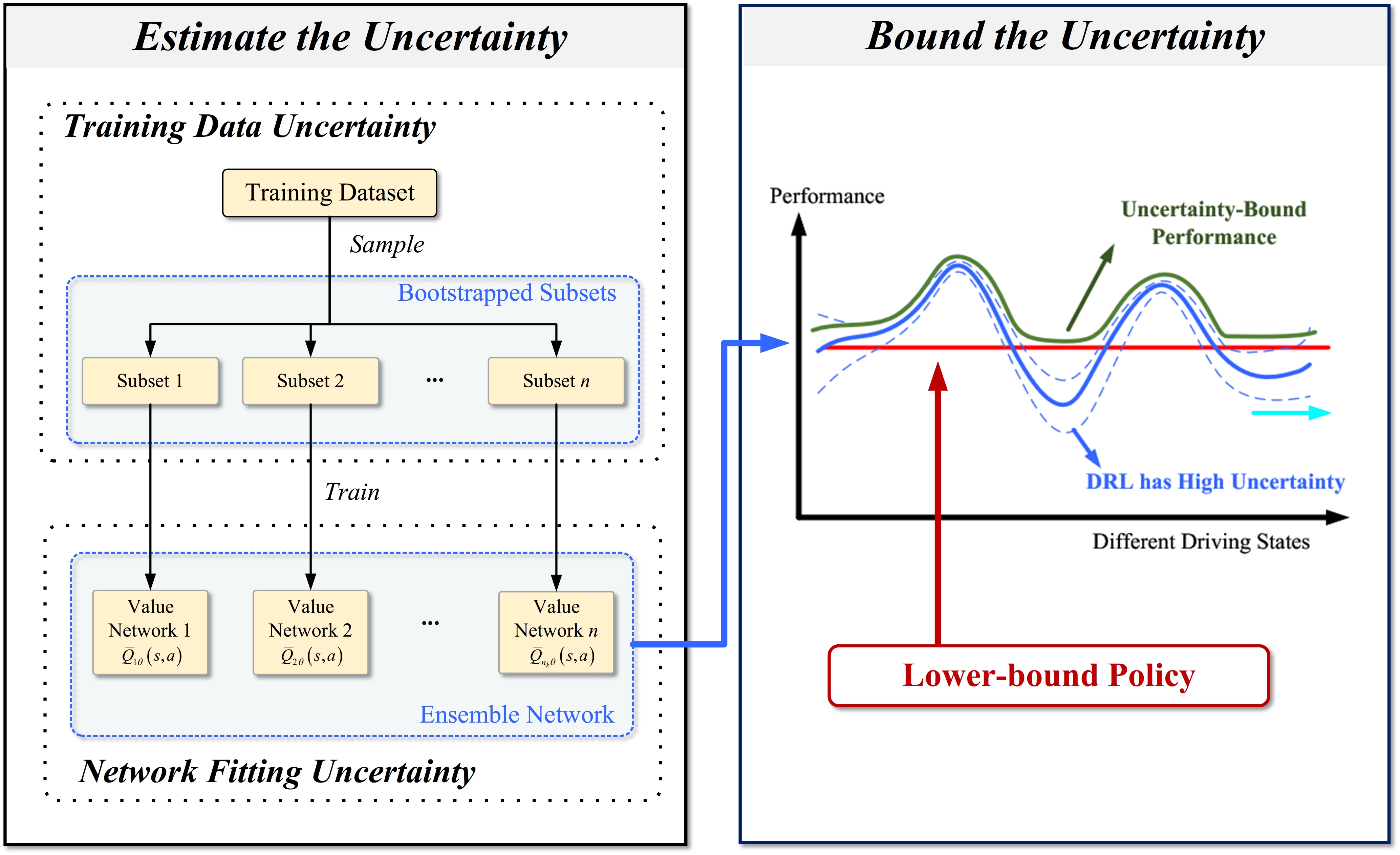}
	\caption{The framework of uncertainty bound reinforcement learning (UBRL). It contains two parts: estimating the uncertainty and bounding the uncertainty. The first component estimates uncertainty by monitoring the policy generation process (highlighted in light yellow), encompassing data collection and network fitting. In the second component, a baseline policy establishes a performance lower bound for the DRL policy.  Consequently, the DRL policy will be protected in instances of high uncertainty or low performance.}
	\label{fig:framework}
\end{figure}

\subsection{Problem Definitions}
{
This study aims to develop a method to identify and protect unreliable decisions of DRL policy. Specifically, there are two primary objectives. First, the method estimates the uncertainty of the DRL policy for an encountered driving case. The uncertainty indicates potential performance degradation due to insufficient training data and network fitting process. Second, the method protects AV using a built performance lower bound when the DRL model has high-uncertain performance. The performance lower bound is constructed  using a baseline policy. The overall UBRL process can be described as follows:

\begin{equation}
a_\text{u} = R_\text{u}(s, \pi_\text{rb}, \pi_\text{rl}, U(s, \pi_\text{rl}, D)) \\
\end{equation}
\begin{equation}
\forall s \in \mathcal{S}, Q_\text{u}(s, a_\text{u}) \geq Q_\text{rb}(s, \pi_\text{rb}(s))
\label{bound_performance_definition}
\end{equation}where $R_\text{u}$ represents  the UBRL process, $a_\text{u}$ denotes  the output action of UBRL agent, and  $s$ is AV's encountered driving case, $\pi_\text{rb}$ is the baseline policy, $\pi_\text{rl}$ is the trained DRL policy, $U(s, \pi_\text{rl}, D)$ is the uncertainty estimation process, $D$ is the training dataset, $Q_\text{u}, Q_\text{rb}$ are the value functions of the UBRL policy and baseline policy, respectively. The UBRL agent builds a performance lower bound using the baseline policy to protect AV when the DRL model has high uncertainty and may have low performance. In other words, the UBRL agent should always outperform the baseline policy $\pi_\text{rb}$.}

\section{Methods}

\subsection{Framework}

The framework of UBRL is shown in Fig.\ref{fig:framework}.  The UBRL method consists of two main components: uncertainty estimation and uncertainty bounding. The uncertainty estimation part tracks the data collection and network fitting process to identify potential DRL model failures. Then, the goal of the uncertainty bounding part is to protect the AV when the DRL model has high uncertainty with a performance lower bound. It should make the UBRL agent always outperform the baseline policy. The methods of each part will be described in detail in the following sections.

\subsection{Data Collection Uncertainty Estimation}

This section introduces  a bootstrap-based method to estimate the uncertainty of the DRL policy arising from insufficient training data. 

\subsubsection{Uncertainty Representation}

As described above, the training process of a typical DRL agent can be described as:
\begin{equation}
\tilde{Q}(s, a )\leftarrow Q_\theta\left(s, a\right) = f(D)
\label{Q_estimate}
\end{equation}
where $D$ denotes the training dataset, and $f$ denotes the training function for network fitting. $Q_\theta(s,a)$ is the learned value function and $\tilde{Q}(s, a )$ denotes the true value. 
The equation describes the assumption that: with a properly-designed training function $f$, the $Q_\theta\left(s, a\right)$ will approach the $\tilde{Q}(s, a)$ when there is sufficient training data \cite{sutton2018reinforcement}. In other words, the trained value function $Q_\theta\left(s, a\right)$ may not be accurate in data-sparse cases, leading to potential low DRL performance.

The uncertainty arising from insufficient training data is represented by a distribution $P\left(\tilde{Q}(s, a) | D\right)$.
The distribution represents  the statistical probability of true value $\tilde{Q}(s, a)$ given the current data $D$, i.e., the confidence of value function estimation. Intuitively, the confidence will be high  with sufficient data, i.e., the probability distribution will be concentrated near the true value. In this case, the estimated value function has a higher probability of accurately approaching true value. At the same time, the confidence of value estimation will be low in data-sparse cases, indicating high DRL model uncertainty due to insufficient training data.
{
The uncertainty arising from insufficient training data is represented as an uncertainty set $U_\text{b}$:

\begin{equation}
U_\text{b} = \{Q_{1b}(s,a), Q_{2b}(s,a), ... Q_{n_\text{e}b}(s,a) \}
\end{equation}
where $Q_{nb}(s,a)$ is defined as a value that is sampled independent and identically distributed (iid) from the distribution $P\left(\tilde{Q}(s, a) |  D\right)$. In this way, the uncertainty set $U_\text{b}$ indicates the DRL uncertainty due to insufficient training data.}

\subsubsection{Uncertainty Estimation based on Bootstrap Principle}

The uncertainty set $U_\text{b}$ is estimated based on bootstrap  principle \cite{efron1992bootstrap}. The bootstrap  principle is to approximate a population distribution by a sample distribution. A classical bootstrap  method takes a dataset $D_\text{b}$ and an estimator $\phi$ as input. 
Commonly, the $\phi$ can estimate a specific variable (e.g., Q-value) from the dataset as $\phi(D_\text{b})$. To obtain the confidence of such an estimation, a bootstrap  distribution is essential but will not be explicitly calculated. 
Instead, several subsets $\overline D_\text{b1}, \overline D_\text{b2}, ... \overline D_\text{bn}$ are  sampled uniformly with replacement from $D_\text{b}$. That is, each subset contains the same amount of data units as the original $D_\text{b}$, but each trajectory in these subsets is uniformly sampled from $D_\text{b}$. Then a sample from the distribution can be calculated using $\phi(\overline  D_\text{bn})$.

Following bootstrap  principle, we sample $n_e$ subsets from the DRL's training dataset $D$. A bootstrap  sample $Q_{nb}(s,a)$ is calculated from a subset $\overline D_n$ with the training function $f$, following Eq. \eqref{Q_estimate}. According to bootstrap  principle, the bootstrap  distribution can be considered as an estimation of the $P\left(\tilde{Q}(s, a) | D\right)$. Thus, the estimated values $Q_{nb}(s,a)$ can be considered as samples from the distribution  $P\left(\tilde{Q}(s, a) |  D\right)$.
In this way, the uncertainty set $U_\text{b}$ contains $n_\text{e}$ bootstrap  values and represents the DRL uncertainty arising from insufficient training data. 
The whole process described is defined as the bootstrap  estimator.

\subsection{Network Fitting Uncertainty Estimation}

This section will introduce an ensemble network method to estimate the uncertainty of DRL policy due to neural network fitting error. 
 
\subsubsection{Uncertainty Representation}

An ensemble of network $U_{\theta}$ is designed to represent uncertainty from the network fitting process, which contains $n_\text{e}$ parallel value networks:

\begin{equation}
U_{\theta}=\{\overline Q_{1\theta}\left(s, a\right),\overline Q_{2\theta}\left(s, a\right) , ... \overline Q_{n_\text{e}\theta}\left(s, a\right)\}
\end{equation}
where  $\overline Q_{n\theta}\left(s, a\right)$  denotes a value network. The structure of the ensemble network is shown in Fig.\ref{fig:framework}. Each value network in $U_{\theta}$ is designed to have the same structure and be randomly initialized as:
\begin{equation}
U_\text{i}=\{Q_{1i}(s,a), Q_{2i}(s,a), ... Q_{n_\text{e}i}(s,a)\}
\end{equation}
where $Q_{1i}, Q_{2i}, ... Q_{n_\text{e}i}$ are the initial values of the $n_\text{e}$ value networks. 

The uncertainty due to network fitting error is captured through the outputs of $n_e$ networks. To achieve that, the initial values $Q_{1i}, Q_{2i}, ... Q_{n_ei}$ are designed to be sampled from a uniform distribution. In this way, the outputs of different networks will have high variance at the beginning of training. Following Eq. \eqref{Q_estimate}, the outputs in $U_{\theta}$ will finally approach the same value (true Q-value) if the training function $f$ is properly designed and the network is well-fitted. Otherwise, the outputs may have a higher variance affected by their initial values, indicating network fitting uncertainty. In this way, the values of  $U_{\theta}$ represent the uncertainty due to network fitting error.

\subsubsection{Uncertainty Estimation and Ensemble Network Training Process}

To capture the uncertainty arising from both insufficient data and network fitting error, the $n_e$ Q-networks in $U_{\theta}$ will be trained with bootstrap  subsets $\overline D_1, \overline D_2, ... \overline D_{n_\text{e}}$  correspondingly using $f$:
\begin{equation}
\overline Q_{n\theta}\left(s, a\right) = f(D_{n})
\label{subset_training}
\end{equation}
The networks in the ensemble network $U_{\theta}$ are then affected by both the training dataset and the network fitting process. Ideally, the values in $U_{\theta}$ will approach those in $U_\text{b}$ when there's no uncertainty from the network fitting process $f$:

\begin{equation}
Q_{nb}(s,a) \leftarrow \overline Q_{n\theta}\left(s, a\right) = f(D_{n})
\end{equation}

Considering the network fitting process, the values in the ensemble network $U_{\theta}$ should approach gradually from the initial values in $U_i$ (represents a uniform distribution) to the bootstrap  estimated values in $U_\text{b}$ (represents $P\left(\tilde{Q}(s, a) |  D\right)$) :

\begin{equation}
\overline Q_{n\theta}\left(s, a\right) := \overline Q_{n\theta}\left(s, a\right) - \alpha\nabla \mathcal{L}(\overline Q_{n\theta}\left(s, a\right),Q_{nb}(s,a))
\end{equation}
where $\alpha$ is the learning rate, $\mathcal{L}$ is the loss function determined by current network output $\overline Q_{n\theta}\left(s, a\right)$  and the bootstrap  target value $Q_{n{b}}(s,a)$.

In this way, the values in $U_{\theta}$ capture the uncertainty arising both from the training data (determined by $U_\text{b}$) and the network fitting process (determined by $U_i$ and the training progress).  The values will have high variance values when training is inadequate (affected more by $U_i$ ).

\subsection{Performance Lower Bound for DRL}

\subsubsection{Definition  of Performance Lower Bound}

The performance lower bound in the UBRL method is built based on a baseline driving policy $\pi_\text{rb}$. {Following Eq.\eqref{bound_performance_definition}, the UBRL agent is bound by the baseline policy means that the UBRL agent should always outperform the baseline policy during driving. 
Based on the estimated DRL model uncertainty, the probability that the DRL policy can outperform the baseline policy at state $ s$ is:

\begin{equation}
\begin{array}{l}
\forall s \in \mathbb{S}: \\
P_\text{o}\left(s\right)=P\left(Q_\text{rl}\left(s, \pi_\text{rl}\left(s\right)\right) \geq Q_\text{rb}\left(s, \pi_\text{rb}\left(s\right)\right)\right)
\label{confidence}
\end{array}
\end{equation}
where $\pi_\text{rl}$, $\pi_\text{rb}$ denote the DRL and baseline policy respectively.}

\subsubsection{Performance Lower Bound Calculation}

The probability is approximated using the estimated uncertainty set $U_{\theta}$:

\begin{equation}
\begin{array}{l}
\forall s \in \mathbb{S}: \\
P_\text{o}\left(s\right)=\frac{\sum_{n=1}^{n_\text{e}} {p}_n\left(s\right)}{n_\text{e}}
\end{array}
\end{equation}
\begin{equation}
p_{n}(s)=\left\{
    \begin{aligned}
    1 & , & \overline Q_{n\theta}\left(s, \pi_\text{rl}\left(s\right)\right) > \overline Q_{n\theta}\left(s, \pi_\text{rb}\left(s\right)\right), \\
    0 & , & \overline Q_{n\theta}\left(s, \pi_\text{rl}\left(s\right)\right) < \overline Q_{n\theta}\left(s, \pi_\text{rb}\left(s\right)\right).
    \end{aligned}
\right.
\end{equation}
where $Q_{n\theta}\in U_{\theta}$ is a value function in the ensemble network. To design the conditions of using the DRL policy in the current state $s$, we consider the Q-value as well as the estimated uncertainty:

\begin{equation}
\begin{array}{l}
\mathbb{E}\left(Q_\text{rl}\left(s, \pi_{r l}\left(s\right)\right)\right)-\mathbb{E}\left(Q_\text{rb}\left(s, \pi_\text{rb}\left(s\right)\right)\right) \geq 0 \\
\mathbb{E}\left(Q_\text{rl}\left(s, \pi_{r l}\left(s\right)\right)\right) = \frac{\sum_{n=1}^{n_\text{e}} \overline Q_{n\theta}\left(s, \pi_\text{rl}\left(s\right)\right)}{n_\text{e}}\\
\mathbb{E}\left(Q_\text{rb}\left(s, \pi_\text{rb}\left(s\right)\right)\right) = \frac{\sum_{n=1}^{n_\text{e}} \overline Q_{n\theta}\left(s, \pi_\text{rb}\left(s\right)\right)}{n_\text{e}}\\
P_\text{o}\left(s\right) > p_{\text {thres }}\\
\end{array}
\label{switch1}
\end{equation}
where $p_{\text {thres }}$ is a manually set parameter. The larger the parameter, the stricter the conditions for using the DRL policy. At the same time, we designed a training count estimator to constraint usage of DRL in extremely data-sparse states:
\begin{equation}
{N_\text{t}}\left(s,\pi_\text{rl}\left(s\right)\right), {N_\text{t}}\left(s,\pi_\text{rb}\left(s\right)\right) \geq n_{\text {thres }}
\label{visit_times_thres}
\end{equation}
where $n_{\text {thres }}$ is a manually set parameter. The $N_\text{t}(s,a)$ is defined as the number of times the state-action pair $(s,a)$ has been trained during the training process, and $n_{\text {thres }}$ denotes a manually-set threshold. Note that the $N_\text{t}(s,a)$ is not equal to the training data amount in $D$. For example, there may be only one data point in the dataset $D$, but it may be used to fit the network many times.

In this way, the UBRL method determines whether to use DRL by considering the uncertainty of value function estimation. 
When more Q-networks indicate that the DRL policy can outperform the baseline policy, the more confident the framework is in activating DRL. Thus, the baseline policy protects the DRL agent in high-uncertainty states without restricting DRL's learning for better performance.

\subsection{Uncertainty-Bound Reinforcement Learning}

\subsubsection{Network Training and Uncertainty Estimation}

This section describes the implementation of the training function $f$.

During the training process, a prioritized  replay buffer $B$ \cite{schaul2015prioritized} is set up to store the collected driving data and generate bootstrap  subsets. 
More specifically, the collected experience $E_t$  $(s_t, a_t, r_{t}, s_{t+1}, m_t)$ at time $t$ includes a bootstrap  mask $m_t$. The mask decides, for each value function   $\overline Q_{n_e\theta}\left(s, a\right) $, whether or not the experience $E_t$ should be used for training \cite{osband2016deep}. The length of $m_t$ is equal to the number of Q-value networks. For example, if $m_t = {1, 0,1,0, 0}$, then the experience $(s_t, a_t, r_{t}, s_{t+1}, m_t)$ should only be used to train $\overline Q_{1\theta}\left(s, a\right) $ and $\overline Q_{3\theta}\left(s, a\right) $. The components of mask $m_t$  are independently drawn from a Bernoulli distribution. In this way, different Q-networks will be trained with corresponding subsets (Eq. \eqref{subset_training}).

Given an experience $E_t$  $(s_t, a_t, r_{t+1}, s_{t+1}, m_t)$, the learning target of Q-network is estimated using Temporal-Difference (TD) method:
\begin{equation}
\theta_{t+1} \leftarrow \theta_{t}+\alpha\left(y_{t}^{Q}-Q\left(s_{t}, a_{t} ; \theta_{t}\right)\right) \nabla_{\theta} Q\left(s_{t}, a_{t} ; \theta_{t}\right)
\label{q_learning_1}
\end{equation}

\begin{equation}
y_{t}^{Q} \leftarrow r_{t}+\gamma Q\left(s_{t+1}, \pi_\text{rb}(s) ; \theta^{-}\right)
\label{q_learning_2}
\end{equation}
where $\alpha$ is the learning rate and $y_{t}^{Q}$ is the target value  $\theta^{-}$ are target network parameters that are copied from the learning network $\theta^{-}\leftarrow\theta_{t}$ only every $\tau$ time steps and then kept fixed in between updates. 
Instead of updating with $r_{t}+\gamma \max _{a} Q\left(s_{t+1}, a ; \theta^{-}\right) $, the  $ r_{t}+\gamma Q\left(s_{t+1}, \pi_\text{rb}(s) ; \theta^{-}\right)$ is utilized for more stable and reliable network updates.

The UBRL agent is designed first to explore the baseline policy's action. The condition to finish exploring this policy is:

\begin{equation}
\begin{array}{l}
\sigma (\overline{Q}_{n\theta}) < \sigma_{\text{thres}} \\
N_\text{t}(s,\pi_\text{rb}(s)) > n_{\text{thres}}
\label{exploration_condition}
\end{array}
\end{equation}
Then, the UBRL will explore other actions following the $\epsilon$-greedy method widely used in DQN. 

\begin{algorithm}  
\caption{Network training}  
\LinesNumbered  
\KwIn{\text {Value function networks}\text{ with }$n_e$\text { outputs },$B$}  

\For{each episode}
    {
    Obtain initial state from environment $s_0$
    
    Pick $Q_k$ to act using $k\sim \text{Uniform}\{1, . . . ,n_e\}$
            
    \For{step t = 1, . . . until end of episode}  
        {  
        
           \eIf{Condition.\ref{exploration_condition} is met}
          {  
          
              Pick an action according to $a_{t} \in \arg \max _{a} Q_{k}\left(s_{t}, a\right)$.
        
          } {
          
          		Pick an action $a_t = \pi_\text{rb}(s_t)$
        
          }
         Receive $s_{t+1}$  and $r_t$ from environment
            
         Sample bootstrap mask $m_t \sim M$
          
          Add $(s_t, a_t, r_{t}, s_{t+1}, m_t)$ \text{to} $B$
          
          Sample Experience \text{from} $B$ 
          
          Train ensemble networks with Eq. \eqref{q_learning_1} and \eqref{q_learning_2}
    
        }
    }

\end{algorithm}  

\subsubsection{Uncertainty-Bound Action Generation}

This section describes the method to generate the UBRL action during driving.

First, the framework obtains DRL policy from the trained ensemble network $U_{\theta}$. The $a_\text{rl}$ will be obtained through a voting mechanism. That is, for each action in the action space, each value network will judge whether taking that action can outperform $a_\text{rb}$. The actions considered by the most Q-networks to outperform will be regarded as $a_\text{rl}$:

\begin{equation}
a_\text{rl}=\pi_\text{rl}(s)=\arg\max_{a \in \mathbb{A}}  \sum_{n=1}^{n_e} p_n(s, a)
\label{max_confidence}
\end{equation}
\begin{equation}
p_{n}(s, a)=\left\{
    \begin{aligned}
    1 & , & \overline Q_{n\theta}\left(s, a_\text{rl}\right) > \overline Q_{n\theta}\left(s, \pi_\text{rb}\left(s\right)\right), \\
    0 & , & \overline Q_{n\theta}\left(s, a_\text{rl}\right) < \overline Q_{n\theta}\left(s, \pi_\text{rb}\left(s\right)\right).
    \end{aligned}
\right.
\end{equation}

where $Q_{n\theta}\in U_{\theta}$ is a value function in the ensemble network.  In each planning step, the framework will judge whether the DRL action can outperform the lower bound action following Eq. \eqref{confidence}. The specific conditions are shown in Eq. \eqref{switch1} and \eqref{visit_times_thres}, which consider the Q-value difference and its uncertainty.

If the condition is met, the DRL action $a_\text{rl}$ will be the final output $a_\text{u}$. Otherwise, the bound action $a_\text{rb}$ will be used. As a result, the baseline policy will be a performance low bound for the DRL policy.

\begin{algorithm}  
\caption{Uncertainty-bound Reinforcement Learning}  
\LinesNumbered  
\KwIn{Environment state $s$}  
\KwOut{The uncertainty-bound action $a_\text{u}$ } 

Calculate lower bound action $a_\text{rb}$ according to Eq. \eqref{optimal}

Calculate DRL action $a_\text{rl}$ according to Eq. \eqref{max_confidence} 

\eIf{Conditions Eq. \eqref{switch1} and \eqref{visit_times_thres} are met}
{
$a_\text{u} = a_\text{rl}$
}{
$a_\text{u} = a_\text{rb}$

}

\end{algorithm}

\section{Case Study}

\subsection{Experiment Setting}
{
\subsubsection{Evaluation Targets}

The experiment aims to evaluate whether if the proposed UBRL method can identify and protect unreliable AV decisions of a DRL policy.  To this end, we test the proposed method using DRL policies with varying uncertainty levels.  The proposed UBRL method should identify the risky decisions and protect DRL models under different uncertainty levels using a baseline policy. That is, the uncertainty of the DRL model should be estimated during driving. Meanwhile, the UBRL method should guarantee that the DRL model is protected and consistently outperform the baseline policy.

\subsubsection{Generation of DRL Model with Uncertainty }

The DRL models with uncertainty are generated by recording models at various training stages. More specific, the DRL policy will be trained from scratch in a designed complex and stochastic driving scenario. Due to the stochastic nature of the scenario, early-stage DRL models may lack sufficient training data to cover all possibilities or may exhibit incomplete network fitting processes.  Consequently, these models possess higher uncertainty and are more likely to making risky decisions. DRL models at different training stages represent varying levels of model uncertainty.  All the DRL models during training will be recorded to evaluate the performance of the proposed UBRL method.
}

\subsubsection{Test Scenario Design}

\begin{figure}[pt]
	\centering
	\includegraphics[width=8cm]{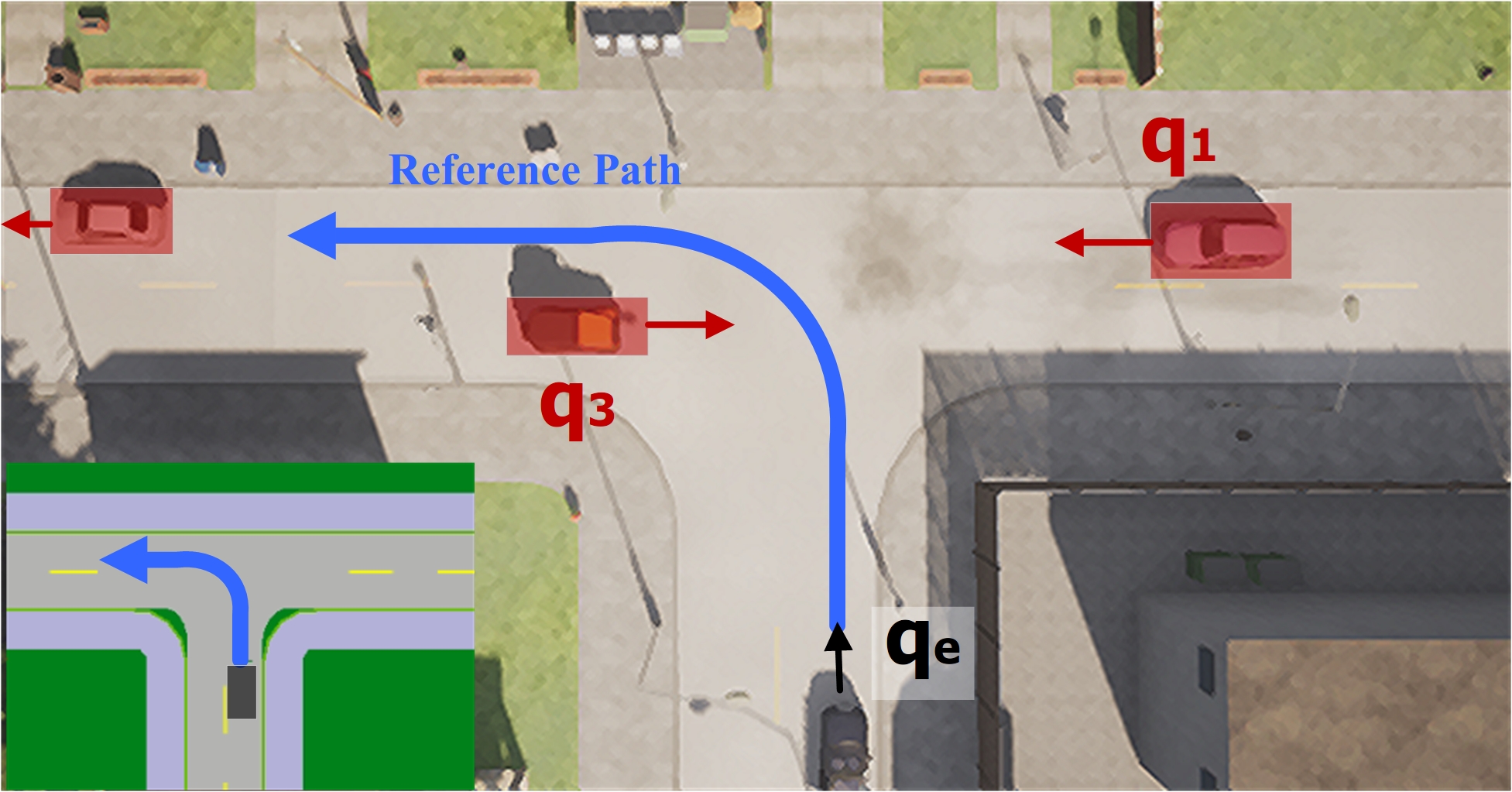}
	\caption{The designed driving scenario. The ego AV (black block) needs to complete an unprotected left-turn driving task at a T-junction (blue arrow). The state space of the problem is also shown in the figure. The states of ego AV ($q_e$) and surrounding vehicles ($q_1, q_2, q_3$,..) from two directions are included. A schematic diagram of the environment is shown in the lower left corner.}
	\label{fig:state_space}
\end{figure}

To train and verify the algorithm, we set up a challenging unprotected left-turn scenario at a T-junction using the CARLA simulator \cite{dosovitskiy2017carla}, as shown in Fig.\ref{fig:state_space}. In this scenario, the ego AV must complete an unprotected left-turn driving task, starting from the lower side of the intersection.  The surrounding vehicles will respond to the ego AV using the driving model built into the CARLA simulator, with randomized parameters to produce stochastic behaviors.

A test episode will end when ego AV finishes the task, collides, or remains stationary for an extended period. Notably, the scenario lacks traffic lights, requiring the ego AV to identify suitable traffic gaps and proceed decisively while considering the stochastic intentions of surrounding vehicles and complex interactions to accomplish the task. The challenging nature of this environment allows for better differentiation between the performance of various driving policies.

\subsubsection{Performance Metric}
In the simulations, the performance metric is defined as the success rate of passing the intersection:

\begin{equation}
{P}_\text{s} =  1 - \frac{N_\text{c} + N_\text{s}}{N_\text{i}}	
\label{collisionrate}
\end{equation}
where $N_\text{c}$ , $N_\text{s}$ and $N_\text{i}$ denote the number of collisions, the number of times the vehicle gets stuck, and the total number of test episodes in the simulations, respectively. During the simulation, a test episode means that the ego AV tries to pass the intersection once. A stuck is defined as when the ego AV keeps stopping for a certain time (i.e., 5s). Successfully passing the intersection, collision, and getting stuck will all be judged as the end of a test episode. To get the performance metrics in the test, the ego AV will make 1000 attempts to pass the intersection to calculate the performance metrics in Eq. \eqref{collisionrate}.

During the test, a driving policy should make decisions for the ego AV in each time step. 
Besides the driving policy part, we obtained other required AV modules, such as trajectory tracking from an open-source platform CLAP \cite{zhong2020clap}. In different simulations, those modules and their parameters are consistent.

\subsection{Uncertainty-Bound Reinforcement Learning Setting}

This section will introduce the agent’s state space, action space, reward function, and training details in the unprotected left-turn scenario. Notably, the settings are applied for the proposed UBRL method and the DRL policy.

\subsubsection{State Space} 
The agent receives state $s$ from the environment in each planning step. The state space is defined as including information about ego AV and surrounding vehicles:

\begin{equation}
s \in \mathcal{S}, s=\left\{q_{e}, q_{1}, q_{2}, q_{3}... q_{m} \right\}   
\end{equation}
where $q_{e}$ denotes the ego AV state, including the vehicle's location, velocity and heading information. Similarly, $q_{m}$ refers to the surrounding traffic states with the same information.
For the designed unprotected left-turn scenario, the surrounding vehicles coming from both directions of the t-junction are included in the state space, as shown in Fig.\ref{fig:state_space}. 

To estimate the training data amount $N_\text{t}(s,a)$, we discretized the continuous state space with step $0.1 \times (\max(s) - \min(s))$ for each dimension $s$. The R-Tree data structure \cite{beckmann1990r} is used for efficient counting.

\subsubsection{Action Space}

Without loss of generality, the action space is designed based on discrete actions. 

\begin{equation}
a \in \mathcal{A}, A = \left(\mathcal{T}_{1}, \mathcal{T}_{2}, \mathcal{T}_{3} \ldots \mathcal{T}_{\mathrm{n}} ,\mathcal{T}_{\mathrm{s}} \right)
\end{equation}where $\mathcal{T}_n$ denotes a pre-designed trajectory. The pre-design trajectories are generated based on lattice planner \cite{werling2010optimal}. The lattice planner is based on a \frenet frame built on the reference path. To generate the trajectories, the planner first samples various trajectory end states $\tau_{1}, \tau_{2}, \tau_{3}, \ldots \tau_{n} $ with different lateral and longitude movements under \frenet frame. 

For each end state $\tau_{n}$, the planner then generates a candidate trajectory  $\mathcal{T}_{\mathrm{n}}$  from quintic polynomials. The quintic polynomial is proved optimal for cost function $C$ in Eq. \eqref{equ:cost1} when giving a start lateral/longitude state $P_{0}=\left[p_{0}, \dot{p}_{0}, \ddot{p}_{0}\right]$ at $t_0$ and $\left[\dot{p}_{1}, \ddot{p}_{1}\right]$ of the end state $P_1$ at time $t_1 = t_0 + T$  under \frenet frame.

\begin{equation}
C=k_\text{j} J_{t}+k_\text{t} g(T)+k_\text{p} h\left(p_{1}\right)
\label{equ:cost1}
\end{equation}
where $J_t$ denotes the time integral of the square of jerk, $g$ and $h$ are arbitrary functions and $k_\text{j}, k_\text{t}, k_\text{p}>0$. As a result, polynomials for different end states are grouped as candidate trajectories $\left(\mathcal{T}_{1}, \mathcal{T}_{2}, \mathcal{T}_{3} \ldots \mathcal{T}_{\mathrm{n}}\right)$.

In the action space, the $\mathcal{T}_{\mathrm{s}}$ denotes a brake trajectory. The shape of this brake trajectory is the same as the candidate trajectory with the lowest cost (without collision checking), and its target velocity is set to zero for each trajectory state.

\begin{figure}[pt]
	\centering
	\includegraphics[width=6cm]{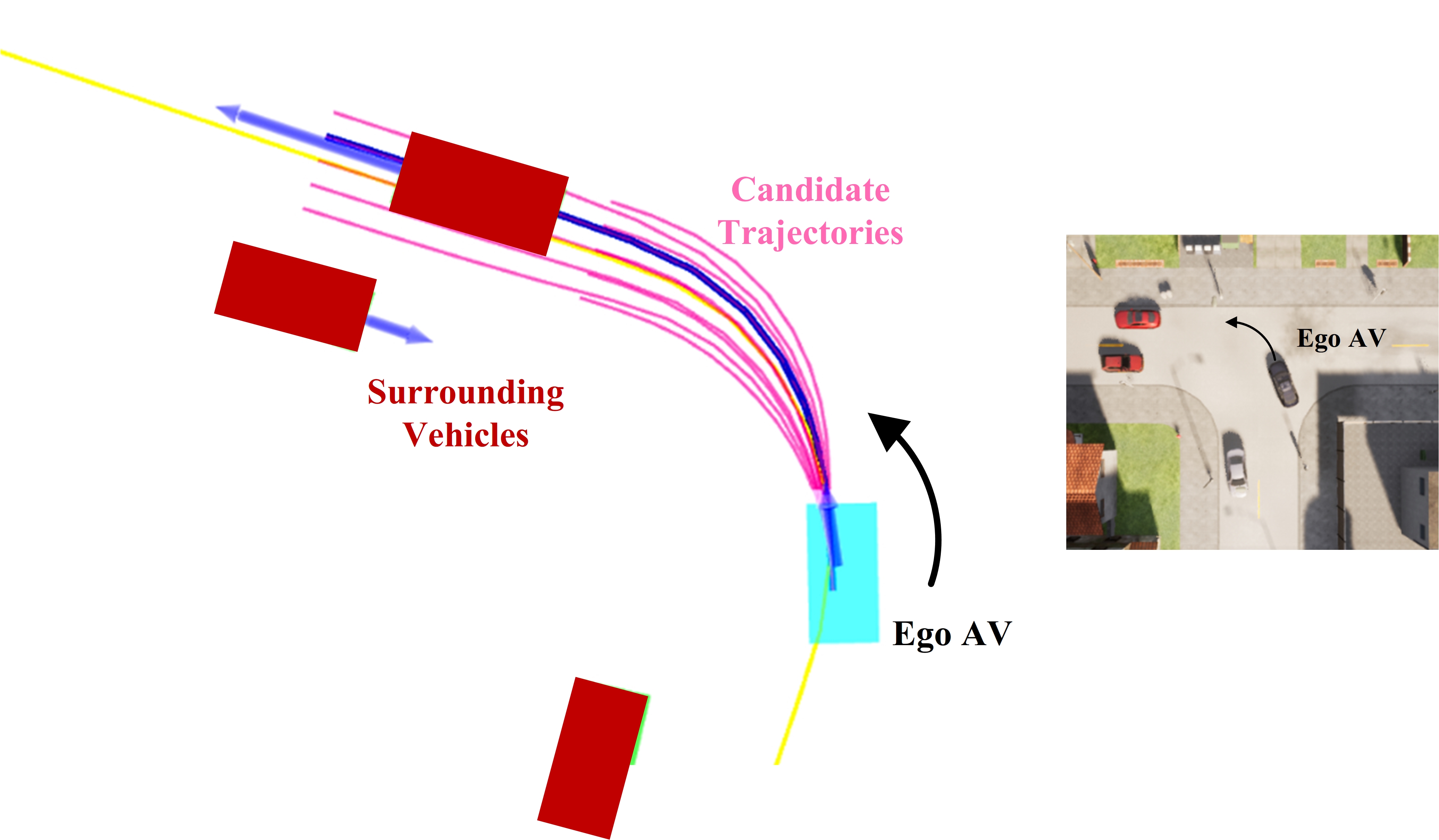}
	\caption{An example of the generated candidate trajectories using the lattice planner. The blue block represents the ego AV, and the red blocks represents the surrounding vehicles. The purple lines represent the candidate trajectories. Additionally, the yellow line denotes the reference path, and the blue line is the final decision trajectory of the planner. The small picture on the right is a top view of the case in the CARLA simulator.}
	\label{fig:ruleplanner}
\end{figure}

In each step, all the new candidate trajectories will be generated by sampling different end states. Each trajectory will keep the state continuity (position, velocity, acceleration) with the chosen trajectory in the last step. Therefore, oscillation planning results (large end-state changes) between two frames will not significantly affect the vehicle's movement. An example of generated trajectories is shown in Fig.\ref{fig:ruleplanner}.

\subsubsection{Reward Function}

The reward function consists of three parts, considering safety and efficiency:

\begin{equation}
r= r_1 + r_2 + r_3
\label{reward_function}
\end{equation}
The first part penalizes collisions:

\begin{equation}
r_{1}=\left\{
    \begin{aligned}
    r_\text{c} & , & collision, \\
    0 & , & else.
    \end{aligned}
\right.
\label{r1}
\end{equation}
The second part rewards the completion of the driving tasks:
\begin{equation}
r_{2}=\left\{
    \begin{aligned}
    r_\text{p} & , & completion, \\
    0 & , & else.
    \end{aligned}
\right.
\label{r2}
\end{equation}
The last part penalizes  over-conservative behavior that causes ego AV to get stuck:
\begin{equation}
r_{3}=\left\{
    \begin{aligned}
    r_\text{s} & , & stuck, \\
    0 & , & else.
    \end{aligned}
\right.
\label{r3}
\end{equation}

\subsubsection{Baseline Policy Design}

In this article,  we used the lattice planner \cite{werling2010optimal} as the baseline policy $\pi_\text{rb}(s)$. This planner can be applied in various scenarios, such as intersections, highways, ramps, etc. The generated trajectory balances comfort, driving efficiency, and safety. The planner includes two main steps: 

First, the planner generates a group of candidate trajectories for space discretization (described in the action space of the previous section ). To sample the candidate trajectories, each trajectory state is defined as $\tau = \{d,\dot{d},b,\dot{b},t\}$, where $d$ and $b$ refer to lateral and longitudinal distances correspondingly. The candidate trajectories are generated by sampling different end states $\tau_{H} =\{d_{H},0,0,\dot{b_{H}},H\}$.

Secondly, the planner selects the safe and optimal trajectory within the generated candidates. The safety of a trajectory  $\mathcal{T}_{\mathrm{n}}$   is guaranteed by collision checking with surrounding objects, as shown in Eq. \eqref{collisioncheck}. The behaviors of the detected objects $(b_{1}, b_{2}, b_{3}... b_{m})$ are predicted by a pre-design model, such as a uniform straight-line model.

\begin{equation}
\phi = {\phi(\mathcal{T}_{\mathrm{n}}, (b_{1}, b_{2}, b_{3}... b_{m}))} = 
\left\{
    \begin{aligned}
    1 & , & collision, \\
    0 & , & else.
    \end{aligned}
\right.
\label{collisioncheck}
\end{equation}
The trajectory with the lowest cost under the premise of safety will be chosen as the planning result:
\begin{equation}
\mathcal{T}_{\text {b }}= \min _{C}\left(\mathcal{T}_{1}, \mathcal{T}_{2}, \mathcal{T}_{3} \ldots \mathcal{T}_{\mathrm{n}}\right) \mid \phi=0
\label{optimal}
\end{equation}
The related parameter settings of the planner are shown in Table \ref{parameter_baseline}. Among them,  $max\_width$ controls the range of $d_{end}$,  $max\_speed$ and $min\_speed$ specifies the range of $\dot{b_{H}}$, and  $H$ is set to control the length of candidate trajectories. Additionally, the number $n$ determines the sampling density. 

In particular, if all candidate trajectories fail to pass the collision checking, the planner will output a braking trajectory $\mathcal{T}_{\text {s}}$ to help the ego AV stop immediately (described in the action space of the previous section). 

Notably, the baseline policy can be any existing rule-based AV algorithm but not explored in this work. The planner's chosen action (trajectory) will be converted into the world coordinate systems. The ego AV's control module (PID velocity controller and Pure Pursuit steering controller) will track the planned trajectory.

\begin{table}
\caption{Parameters of Baseline Policy}
\begin{center}
\begin{tabular}{ccc}
\toprule  
Parameters& Symbol& Value\\
\midrule  
Sample candidate trajectories number & $n$ & 7\\
Lateral sample limits  & $max\_width$ & 2.0m\\
Maximum sampling speed & $max\_speed$ & 30km/h\\
Minimum sampling speed & $min\_speed$ & 0km/h\\
Candidate trajectories length & $t_{end}$ & 6s\\
Comfort weight & $k_j$ & 0.1 \\
Efficiency weight & $k_t$ & 0.1 \\
Follow weight & $k_p$ & 1.0 \\
Planning frequency  & -- & 10 Hz \\
\bottomrule 
\end{tabular}
\end{center}
\label{parameter_baseline}       
\end{table}

\subsubsection{Neural Network Structure and Training Setting}

The ensemble network structure is shown in Fig.\ref{fig:framework}. The network contains ten separate Q-networks ($n_e = 10$), each containing four fully connected layers separately. Each network predicts a Q-value $\overline Q_{k\theta}\left(s, a_{\pi}\right) $. The weight of the fully connected layers in the heads is initialized randomly using a uniform distribution.

During the training process, we combined several practical improvements of the DQN algorithm, including Double DQN \cite{van2016deep} and Priority Replay Buffer \cite{schaul2015prioritized}. Similar to Double DQN, each Q-network in the ensemble network contains a target network ${\theta}_{t}^{\prime}$ that is used to calculate the updated value:

\begin{equation}
y_{t}^{Q} \leftarrow r_{t}+\gamma Q(s_{t+1}, a_b ; \theta_{t}^{\prime})
\label{ddq_learning}
\end{equation}
The prioritized  replay buffer directly replaces the replay buffer in DQN to increase the sampling frequency of dangerous cases. Relevant parameters of UBRL method are shown in Table \ref{parameter_drl}.

\begin{table}
\caption{Parameters of the proposed UBRL}
\begin{center}
\begin{tabular}{ccc}
\toprule  
Parameters& Symbol& Value\\
\midrule  
Collision penalty & $r_\text{c}$ & 0 \\
Task success reward & $r_p$ & 1 \\
Stuck penalty & $r_s$ & 0 \\
Learning rate & $\alpha$ & 5e-4 \\
Discount factor& $\gamma$ & 0.995\\
Batch size & -- & 64 \\
Planning frequency & -- & 10 Hz\\
Heads number & $n_e$ & 10\\
Exploration var parameter & $k_e$ & 0.01\\
bootstrap  data share probability & -- & 0.8 \\
Threshold of uncertainty estimation & $\sigma_{\text{thres}}$ & 0.05 \\

\bottomrule
\end{tabular}
\end{center}
\label{parameter_drl}       

\end{table}

\subsection{Test Results}

We trained the UBRL model and DRL model in the designed driving scenario. The models have collected over 300,000 frames of driving data (approximately 30,000 training episodes). During this process, we recorded all the intermediate models for evaluation. The results are introduced in this section. 

\subsubsection{Uncertainty Estimation Performance of UBRL}

\begin{figure}[pt]
	\centering
	\includegraphics[width=0.95\linewidth]{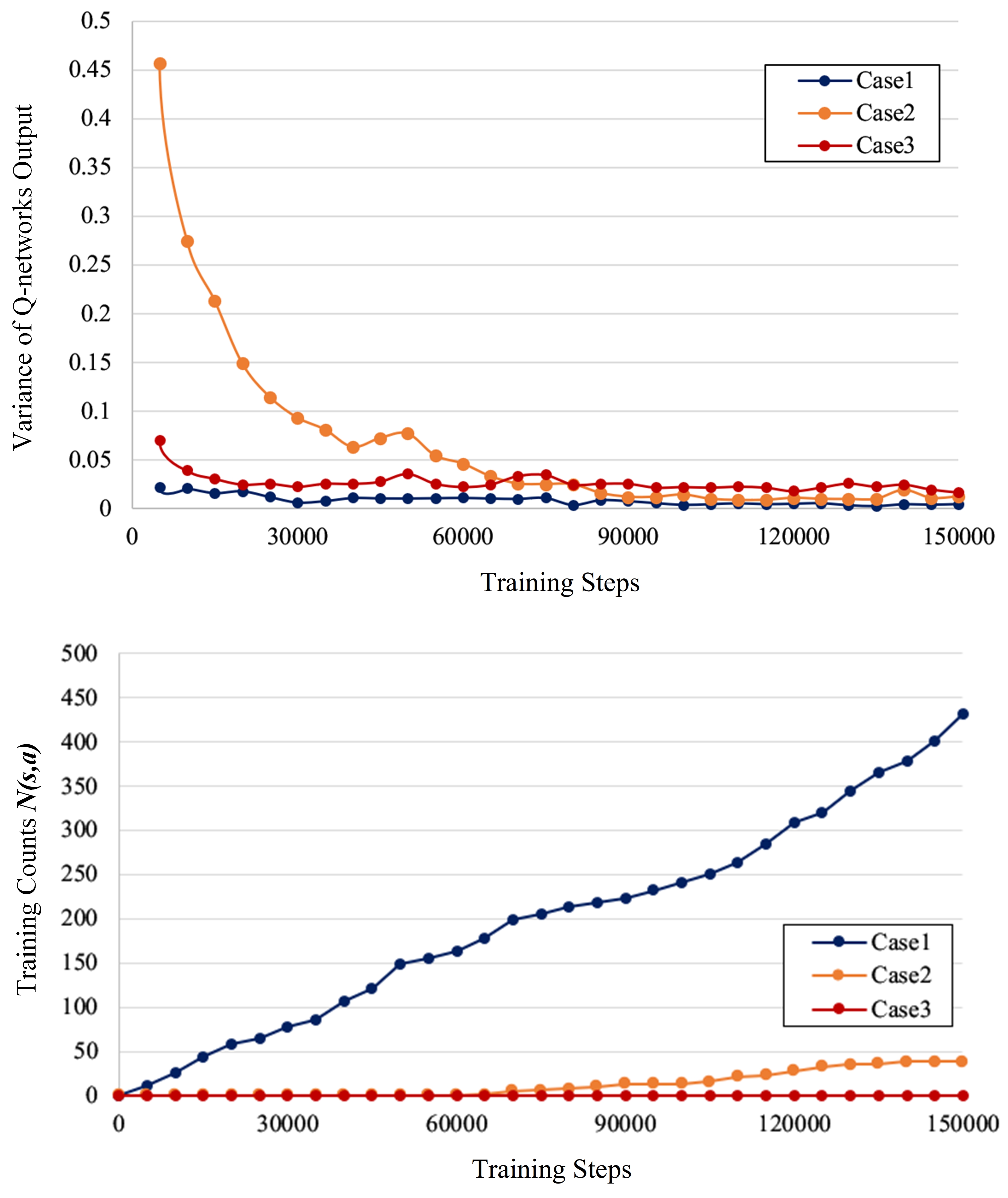}
	\caption{The uncertainty estimation of three example cases. The upper part shows the variance of different Q-networks in the ensemble structure. The lower part shows each case's training data amount $N_\text{t}(s,a)$.
}
	\label{fig:bootstrap_result_1}
\end{figure}

\begin{figure}[pt]
	\centering
	\includegraphics[width=0.95\linewidth]{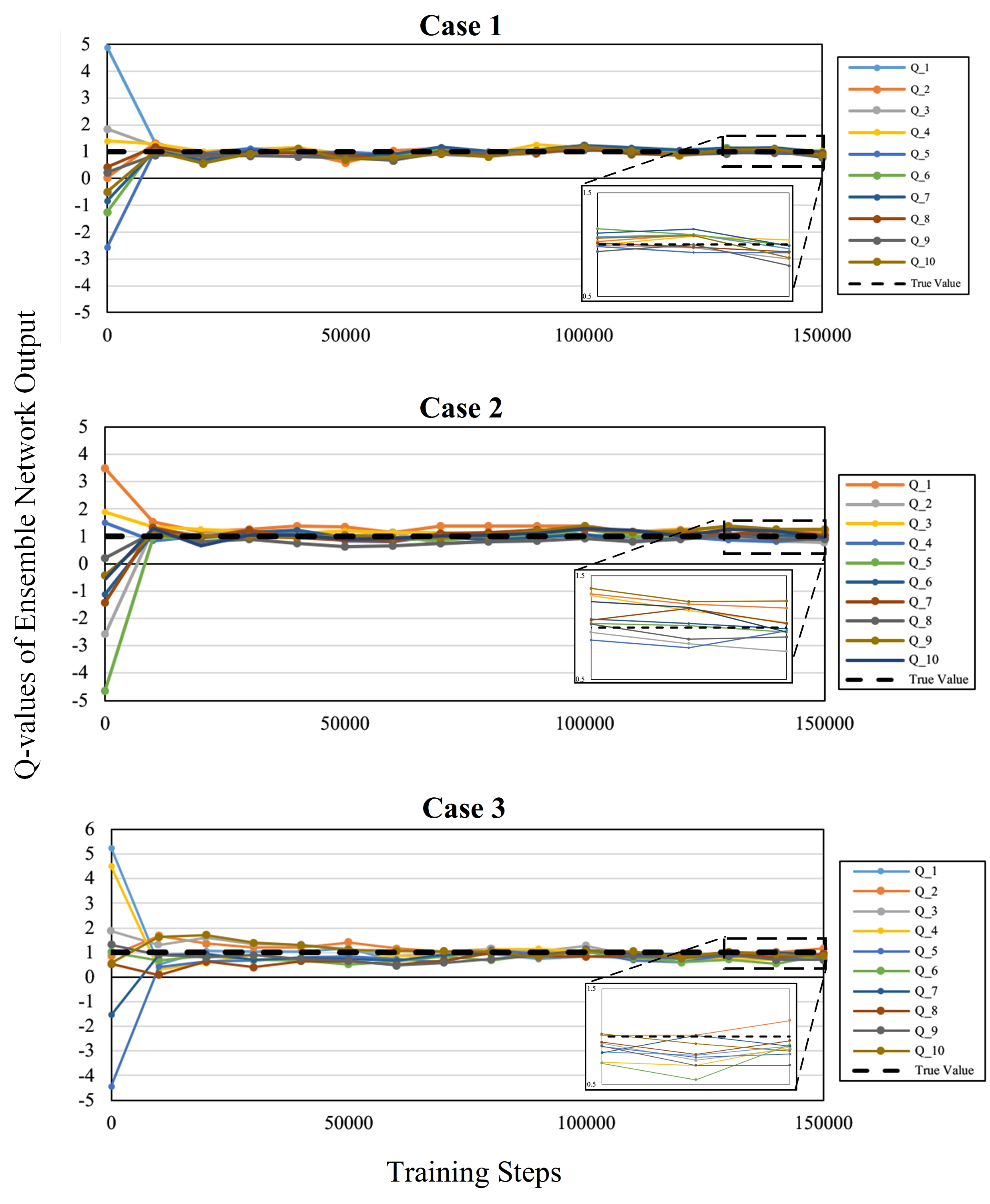}
	\caption{The recorded $n_e$ Q-values predicted by the ensemble network during the training process for the three example cases. The true Q-values in the cases is represented with black dotted lines.
}
	\label{fig:bootstrap_result_2}
\end{figure}

We choose three representative example cases (state-action pairs) to show the uncertainty estimation performance during driving. The results are shown in Fig.\ref{fig:bootstrap_result_1} and \ref{fig:bootstrap_result_2}. The states are chosen from terminal states during driving since the true Q-value of these cases is easy to calculate accurately:
\begin{equation}
Q(s,a) = r_\text{T}
\end{equation}where $r_\text{T}$ is the terminal reward for the terminal state.

Fig.\ref{fig:bootstrap_result_1} shows uncertainty estimation results of the three example cases. The upper part shows the variance of different Q-networks in the ensemble structure. The lower part shows the training counts $N_\text{t}(s,a)$ of different cases. 

Case 1 is a typical ``low-uncertainty'' case. The ensemble network outputs have small variance after training (0.005). The network has been trained 431 times in this case. The UBRL method considers it as a ``reliable" case that the DRL model's performance is guaranteed.
The DRL model has higher uncertainty in case 2 than in case 1. This is because the agent collects limited data of case 2: 0 until 60,000 steps and 38 after 150,000 steps. In this case, the Q-values converge near the true value but have a higher variance (0.013). The DRL model has the highest uncertainty in case 3. This is because the agent collects no data during the whole training process. The ensemble network outputs the highest variance (0.017) in this case, and the UBRL method considers it an ``unreliable'' case.

Fig.\ref{fig:bootstrap_result_2} shows the output of $n_e$ Q-values predicted by the ensemble network during the training process. The Q-values in the ensemble network converge more (to the true value) in the  ``low-uncertainty'' case 1. In case 2 and case 3, the values have higher variance. The true Q-value is closed to the Q-networks' outputs in both cases. 
The three examples in this section illustrate how the UBRL method estimates the DRL model's uncertainty through the outputs of the ensemble network.

\subsubsection{Uncertainty-Bound performance of UBRL}

\begin{figure}[pt]
	\centering
	\includegraphics[width=9cm]{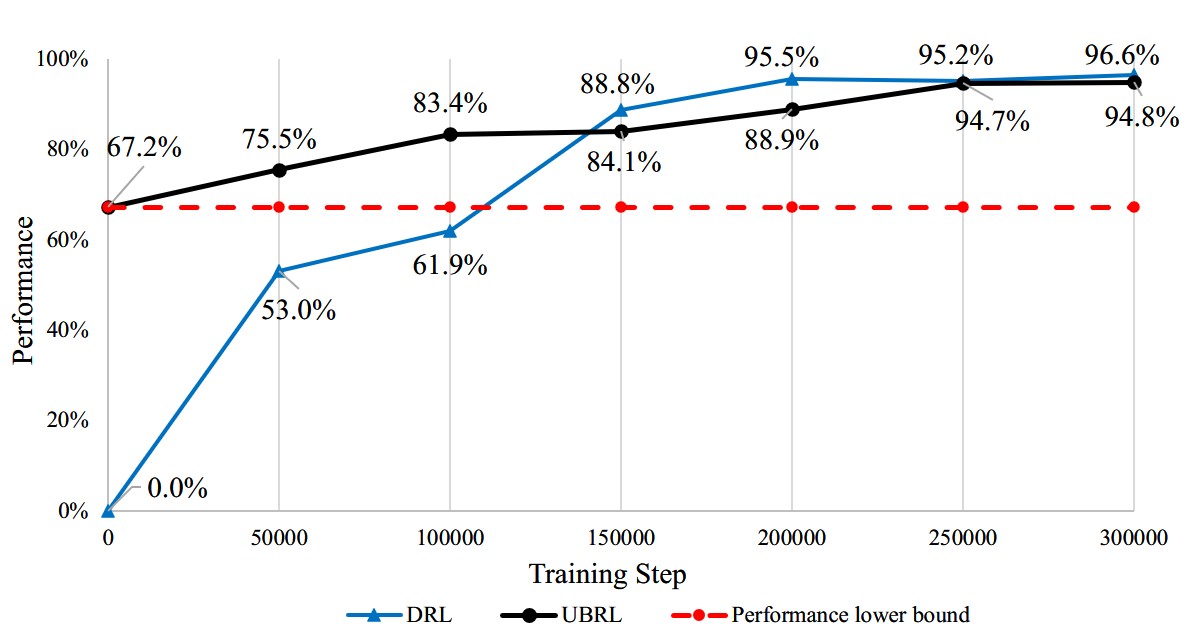}
	\caption{The planning success rate of UBRL and DRL at different training stages. The UBRL can outperform the baseline policy regardless of the training stages. Meanwhile, the UBRL agent learns and improves driving performance with more training data.}
	\label{fig:ubp_performance}
\end{figure}

Fig.\ref{fig:ubp_performance} shows the performance of UBRL and DRL policy under different training stages, i.e., the DRL model may have different levels of uncertainty. The DRL policy for comparison behaves badly at the early stages and cannot outperforms the lower bound until approximately 120,000 steps. It finally achieves well performance and converges after about 200,000 training steps. 

The results show that the UBRL can 1) outperform the baseline policy no matter what stage of training and 2) learn and improve driving performance with more training data (from 67.2\% to 94.8\%). 
On the one hand, the UBRL method can well identify those unreliable decisions of DRL policy through uncertainty estimation. Moreover, the baseline policy is only activated when the DRL policy is unreliable (Eq. \eqref{switch1} and \eqref{visit_times_thres}). 
On the other hand, the DRL actions that can outperform the lower bound are kept. Thus the UBRL can outperform the baseline policy (at any training stage). As long as there is a case where DRL is activated, it will help improve the overall performance of UBRL. Secondly, with gradually increased training data and network fitting process, the UBRL will find the DRL policy reliable and outperform the baseline policy in more and more states. Then it will activate the DRL more for better performance. Thus the UBRL performance can improve with more training data. 

The DRL policy for comparison cannot outperform the lower bound until about 120,000 steps. Thus it is hazardous to use the DRL policy before that. Notably, the amount of data (120,000 frames) is only enough for the scenario we designed, and the amount of data required in real-world driving may far exceed  \cite{kalra2016driving}. Thus a DRL policy may be unreliable for a long period (and we can not know or explain when it will fail). The proposed method can help AVs have self-learning ability while maintaining at least the performance of an existing baseline policy. At the same time, our method helps predict when the DRL policy may make unreliable decisions and help explain why it fails, which is helpful to the safety and interpretability of AV software.

\subsubsection{Effect of UBRL Parameters $p_{\text{thres}}$ and $n_{\text{thres}}$}

\begin{figure}[pt]
	\centering
	\includegraphics[width=9cm]{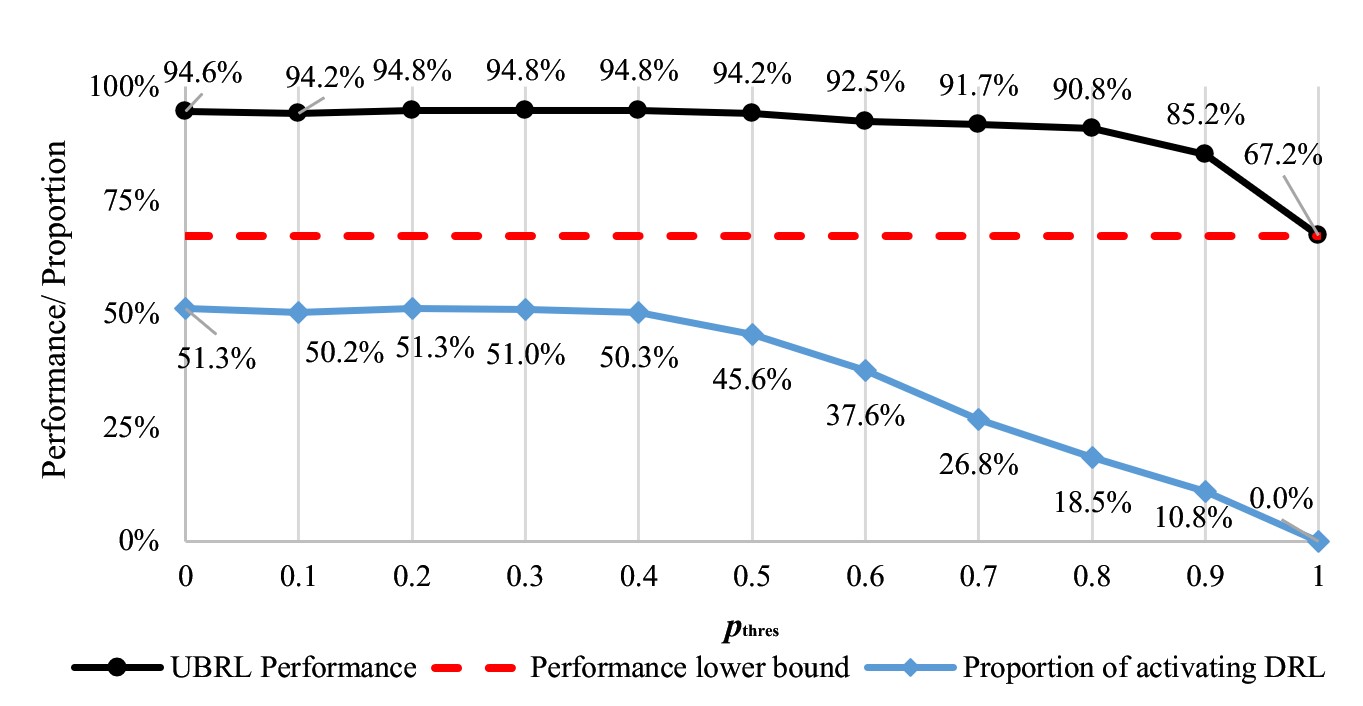}
	\caption{The performance of UBRL at different $p_{\text{thres}}$ values. A higher $p_{\text{thres}}$ value leads to a lower proportion of activating DRL policy, and the performance of UBRL is closer to the baseline policy.}
	\label{fig:pthres}
\end{figure}

Fig.\ref{fig:pthres} shows the performance of UBRL under different $p_{\text{thres}}$ (after 300,000 training steps). The proportion of activated DRL indicates the percentage of using DRL action during a whole test process.
According to the UBRL design in Eq. \eqref{switch1}, a larger $p_{\text{thres}}$  leads to fewer DRL activation. It explains the results when $p_{\text{thres}}>0.4$. In this case, the performance of UBRL is also reduced with larger $p_{\text{thres}}$, as well-performing DRL actions might be activated less. In the extreme case when  $p_{\text{thres}}$ is set to 1, the UBRL will not activate DRL policy and be equal to the baseline policy. When $p_{\text{thres}}<0.4$, other conditions (such as Eq. \eqref{visit_times_thres}) may play a major role in limiting the activation of the DRL policy, and thus the $p_{\text{thres}}$ has a minor impact.

\begin{figure}[pt]
	\centering
	\includegraphics[width=9cm]{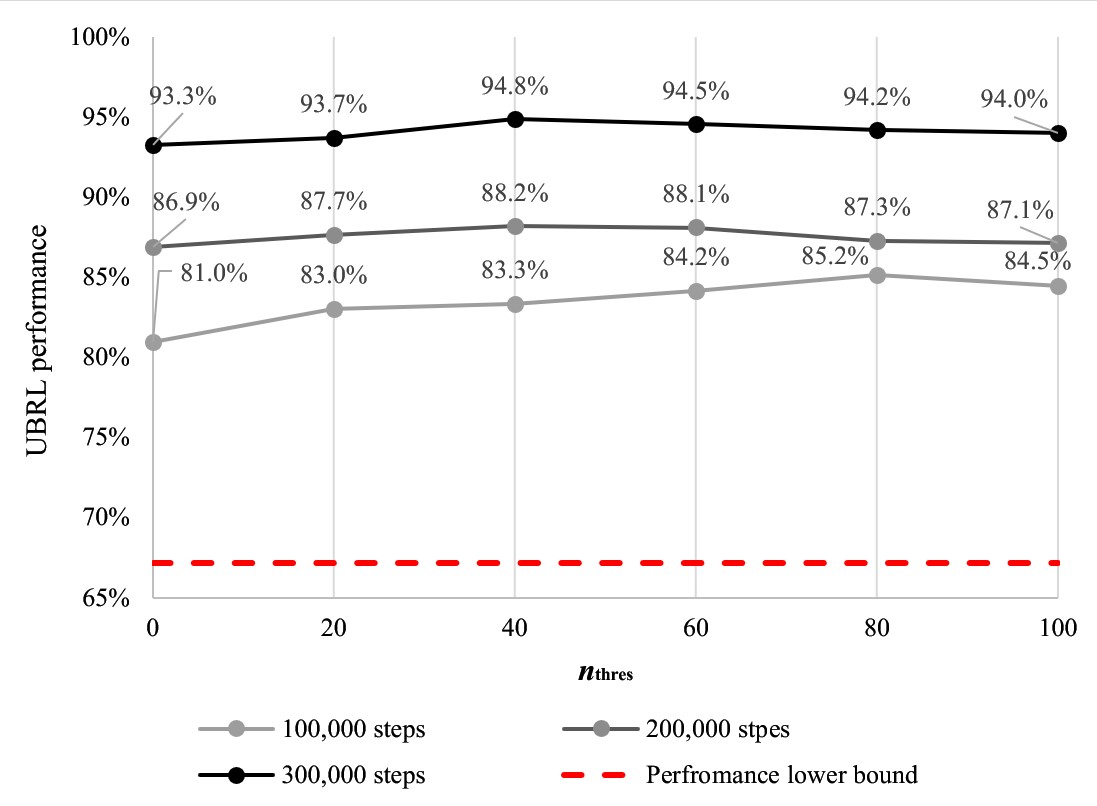}
	\caption{The performance of UBRL under different $n_{\text{thres}}$ values. }
	\label{fig:nthres}
\end{figure}

Fig.\ref{fig:nthres} shows the performance of UBRL under different $n_{\text{thres}}$ (at different training stages). Similar to $p_{\text{thres}}$, the DRL policy will be activated more when $n_{\text{thres}}$  is small. When $n_{\text{thres}}$ is set to 0, the UBRL is equivalent to working without the training counter estimator and might fail to identify some unreliable state-action pairs of DRL policy. The results show that the constraints imposed by the training count estimator (Eq. \eqref{visit_times_thres}) can improve the performance of UBRL. However, when $n_{\text{thres}}$ is set too large (e.g. more than 80), the performance of UBRL may decrease from optimal. 
It might be due to too strict conditions that result in fewer activation of well-perform DRL actions.

\begin{figure}[pt]
	\centering
	\includegraphics[width=1\linewidth]{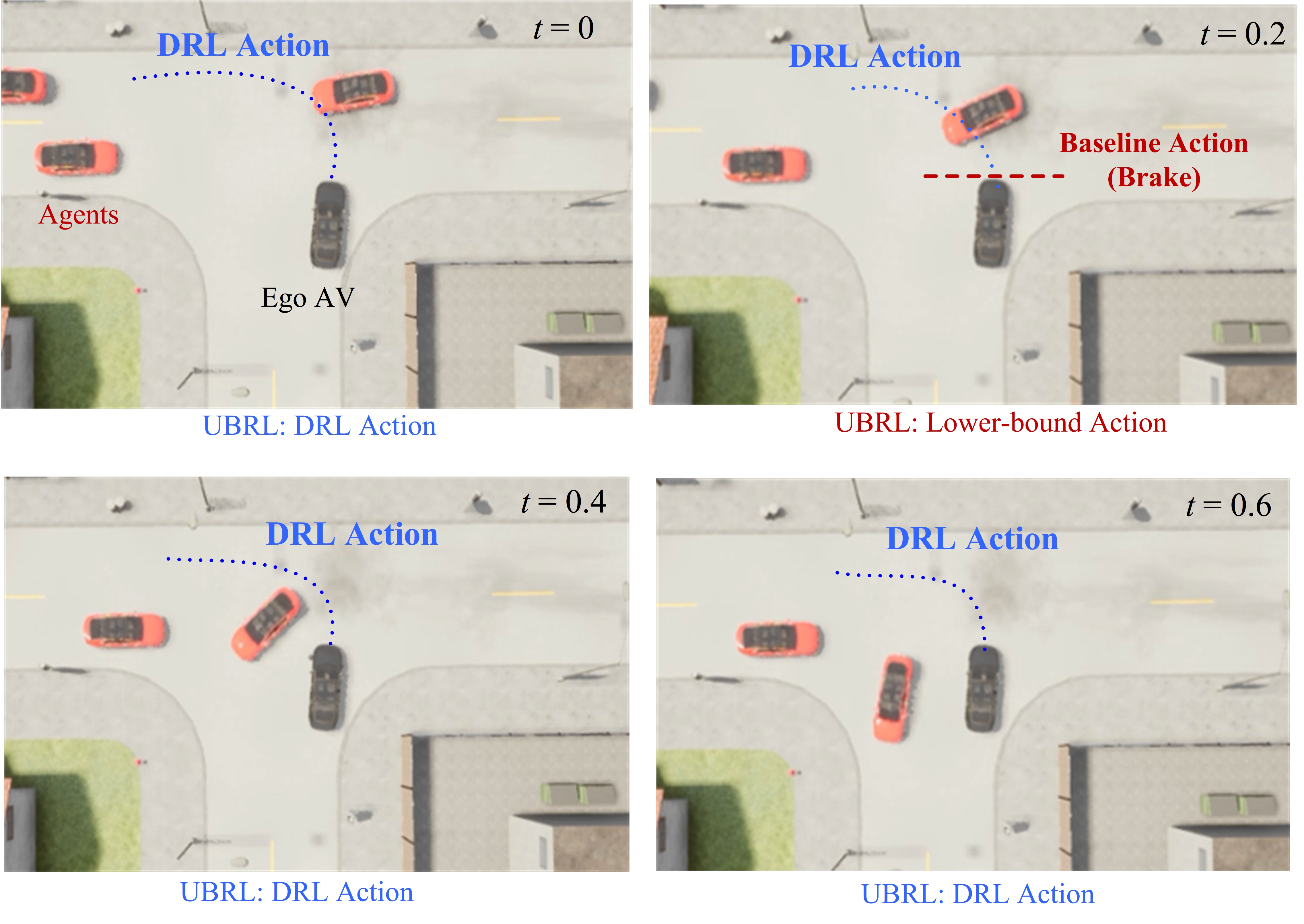}
	\caption{An example cases of how UBRL protects the ego AV from high uncertainty. The four sub-figures show the driving process in time sequence. The UBRL uses the baseline policy when the DRL policy has high uncertainty (when $t=0.2$) and avoids possible collisions.}
	\label{fig:ubp_cases}
\end{figure}

\subsubsection{Example Case of UBRL Driving}

An example case of how UBRL protects the ego AV from the DRL's risky deicision is shown in Fig.\ref{fig:ubp_cases}. More specifically, the four sub-figures show the decisions made by the UBRL method every 0.2 seconds. Among them, when t = 0, the action of DRL is adopted because of low uncertainty. The output variances of the Q-networks are 0.0073. Then, when t=0.2, UBRL finds high uncertainty of DRL action (variance is 0.02784 with 0 training data). Thus, UBRL takes the lower bound action (brakes) instead of the DRL action (continues to pass the intersection). If the vehicle takes the DRL action that drives the vehicle to accelerate (to more than 5m/s), the ego AV may collide with the front left-turn red vehicle. The baseline policy takes a brake and decelerate the vehicle to 0.069 m/s and avoid potential risk. Then, the UBRL takes the DRL action after t=0.6, making the ego AV pass the intersection efficiently.

\section{Conclusion}
This paper proposed an uncertainty-bound reinforcement learning (UBRL) method, which aimed at utilizing a DRL policy with estimated and bounded model uncertainty to achieve more reliable driving performance. The framework first estimates the performance uncertainty of DRL policy by tracking its generation process. Then, three estimators are designed for the uncertainty arising from the data collection distribution and network fitting process. The estimators include bootstrap  estimator, ensemble network estimator, and training count estimator. 
Then a performance lower bound of DRL policy is built considering the uncertainty using a baseline policy. In this way, the UBRL agent should consistently outperform the baseline policy.

The proposed method is evaluated in a challenging unprotected left-turn t-junction scenario in the CARLA simulator. The results show that the UBRL method can reasonably estimate the uncertainty of DRL policy and identify potentially unreliable decisions. Then the built performance lower bound will be activated and protect the DRL policy. The UBRL method can avoid potential catastrophic decisions of DRL model and outperform the baseline policy in all training stages. Meanwhile, the UBRL agent can learn and improve driving performance with more data. The proposed UBRL method provides a potential way to apply the DRL method on AVs reliably.

\section{Acknowledgment}

This work is supported by the National Natural Science Foundation of China (NSFC) (U22A20104, 52102460) and Beijing Municipal Science and Technology Commission (Z221100008122011).
It is also funded by the Tsinghua University-Toyota Joint Center.

\appendices

\ifCLASSOPTIONcaptionsoff
  \newpage
\fi

\bibliographystyle{IEEEtran}
\bibliography{ref.bib}

\begin{IEEEbiography}[{\includegraphics[width=1in,height=1.25in,clip,keepaspectratio]{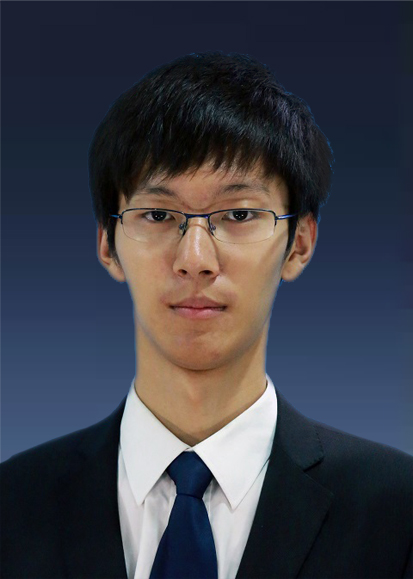}}]{Weitao Zhou} received the B.S. and M.S. in automotive engineering from Beihang University. 

He is currently a Ph.D. student at Tsinghua University. 
His research interests include autonomous driving, reinforcement learning and open-world learning.
\end{IEEEbiography}
\vspace{-20pt}
\begin{IEEEbiography}[{\includegraphics[width=1.1in,height=1.25in,clip,keepaspectratio]{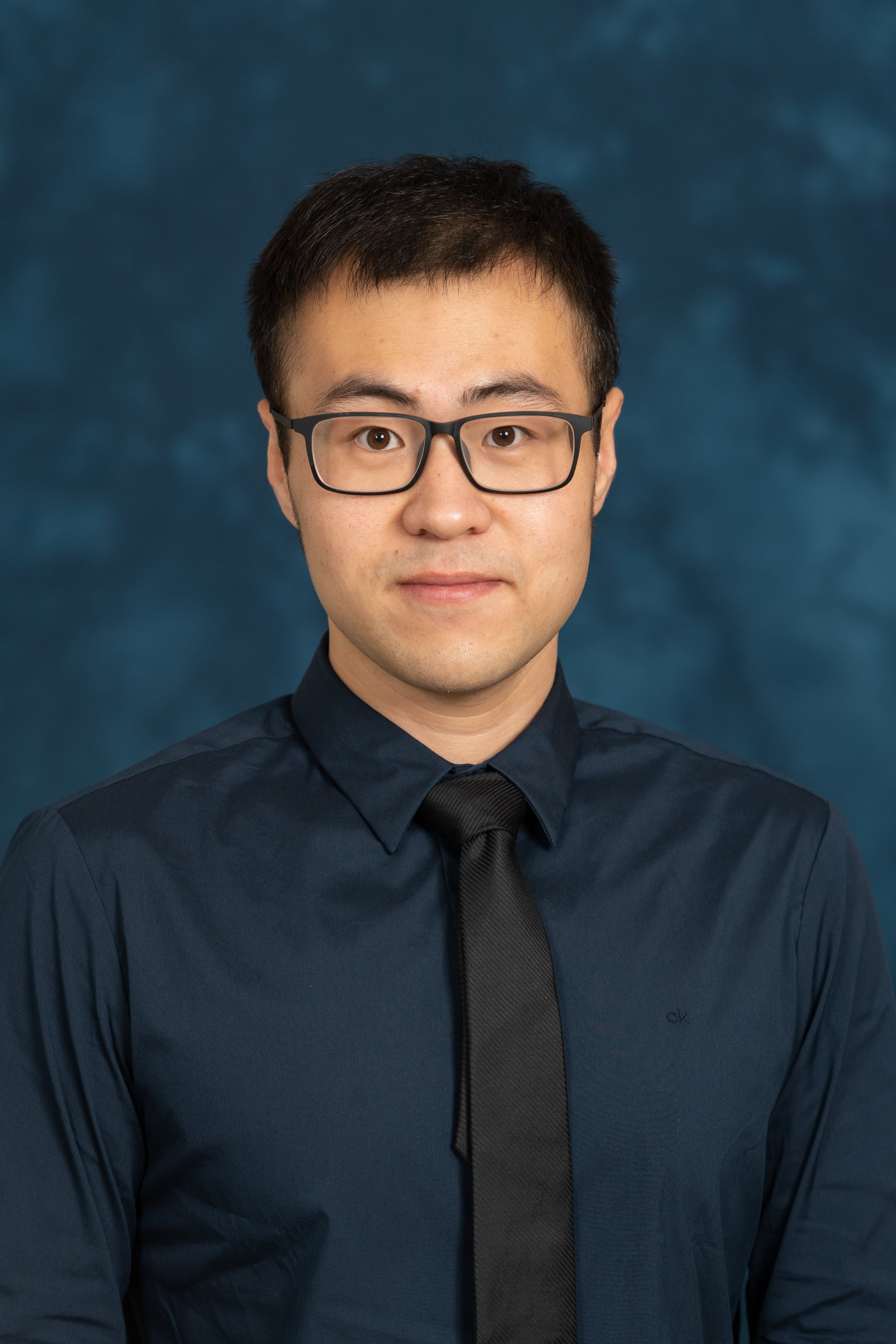}}]{Zhong Cao} received the B.S. and Ph.D. degree in automotive engineering from Tsinghua University in 2015 and 2020, respectively. 

He is currently a postdoc at Tsinghua University, with Shuimu Scholarship. 
His research interests include autonomous vehicle, trustworthy reinforcement learning, long life learning and HD map.
\end{IEEEbiography}
\vspace{-20pt}
\begin{IEEEbiography}[{\includegraphics[width=0.9in,height=1.25in,clip,keepaspectratio]{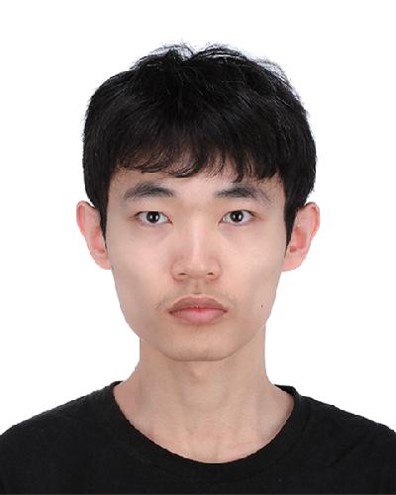}}]{Nanshan Deng} received the B.S. degree in automotive engineering from Tsinghua University in 2018. 

He is currently a Ph.D. student at Tsinghua University. His research interests include autonomous vehicle,  transfer reinforcement learning and meta learning.
\end{IEEEbiography}
\vspace{-20pt}
\begin{IEEEbiography}[{\includegraphics[width=0.9in,height=1.25in,clip,keepaspectratio]{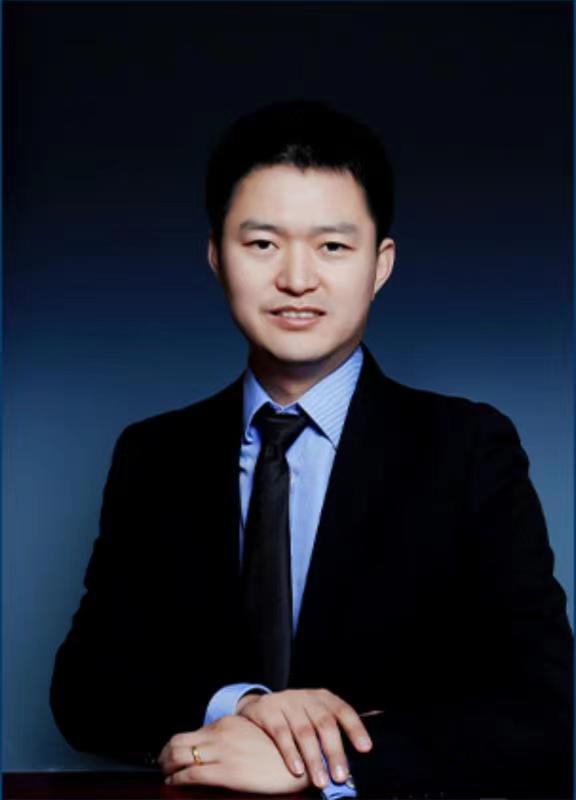}}]{Kun Jiang} received the B.S. degree in mechanical and automation engineering from Shanghai Jiao Tong University, Shanghai, China in 2011. Then he received the Master degree in mechatronics system and the Ph.D. degree in information and systems technologies from University of Technology of Compiègne (UTC), Compiègne, France, in 2013 and 2016, respectively. He is currently an assistant research professor at Tsinghua University, Beijing, China. His research interests include autonomous vehicles, high precision digital map, and sensor fusion.
\end{IEEEbiography}
\vspace{-20pt}
\begin{IEEEbiography}[{\includegraphics[width=1.1in,height=1.25in,clip,keepaspectratio]{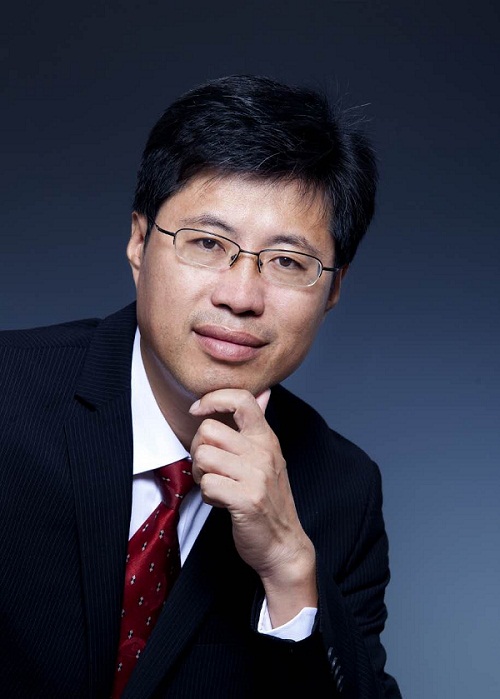}}]{Diange Yang} received the B.S. and Ph.D. degrees in automotive engineering from Tsinghua University, Beijing, China, in 1996 and 2001, respectively. He serves as the director of automotive engineering at Tsinghua university.

He is currently a Professor with the Department of Automotive Engineering, Tsinghua University. His research interests include intelligent transport systems, vehicle electronics, and vehicle noise measurement.

Dr. Yang attended in “Ten Thousand Talent Program” in 2016. He also received the Second Prize from the National Technology Invention Rewards of China in 2010 and the Award for Distinguished Young Science and Technology Talent of the China Automobile Industry in 2011.

\end{IEEEbiography}

\end{document}